%% file: main.tex
\pdfoutput=1
\documentclass[a4paper, 10 pt, conference]{cras}
\usepackage[utf8]{inputenc}
\IEEEoverridecommandlockouts
\usepackage{mathtools}
\usepackage{geometry}
\usepackage{newtxmath,newtxtext}
\usepackage{authblk}
\usepackage{pgfplots}
\usepackage{todonotes}
\usepackage{subfig}
\usepackage{graphicx}
\graphicspath{{./}{../}}
\pgfplotsset{compat=newest}
\pgfplotsset{plot coordinates/math parser=false}

\title{\Large \bf
Compliance-Based Sensor Placement for Force Sensing on a Sensorized Prostate Phantom}

\author{\large Sizhe Tian}
\author{\large Yinoussa Adagolodjo}
\author{\large Jeremie Dequidt}

\vspace{10pt}
\affil{\small\textit{Inria, CNRS, Centrale Lille, UMR 9189 CRIStAL, University of Lille, Lille, France.}}

\begin{document}

\maketitle
\thispagestyle{empty}
\pagestyle{empty}

\section*{INTRODUCTION}

Phantoms are widely used in medical training, 
such as for palpation-based examinations like the Digital Rectal Examination (DRE).
Traditional physical phantoms are passive blocks of silicone that only provide limited feedback to students~\cite{gerling_design_2009}.
In contrast, soft robotic phantoms equipped with sensing provide quantitative evaluation of trainee performance~\cite{escaida_navarro_bio-inspired_2022, hughes_sensorized_2020}.
For this feedback to be meaningful, the phantom must estimate both where and how hard the trainee palpates, 
without restricting contact to pre-defined locations.
The examination context imposes constraints:
vision-based tracking may be occluded by the surrounding anatomy,
and dense sensor arrays compromise the silicone's compliance and complicate fabrication.
These constraints motivate a sparse sensing strategy, where sensor placement directly affects localization accuracy.

Existing methods typically maximize the information content of the configuration
with respect to target modes or deformation
patterns~\cite{manohar_data-driven_2018, zhou_efficient_2018, a_spielberg_co-learning_2021, kim_optimal_2024}.
We build on the QR-based selection strategy of Manohar \textit{et al.}~\cite{manohar_data-driven_2018},
but operate on a compliance matrix derived from FEM simulation and add a weighted greedy
search that prioritizes the clinically relevant DRE contact zone.

\section*{MATERIALS AND METHODS}

The sensor set consists of two modalities: surface displacement markers (e.g., motion capture) 
whose placement we optimize via a compliance matrix, and three internal pneumatic chambers used as intrinsic pressure sensors.

External deformation induces volume changes in each internal chamber,
resulting (via the isothermal Ideal Gas Law, $PV = \text{const}$) in measurable pressure variations.
Chamber pressures thus provide a global volume-derived signal
that complements the sparse local displacements of the surface markers.

\begin{figure}[!t]
    \centering
    \fontsize{6}{8}\selectfont
    \def\svgwidth{0.9\columnwidth}
    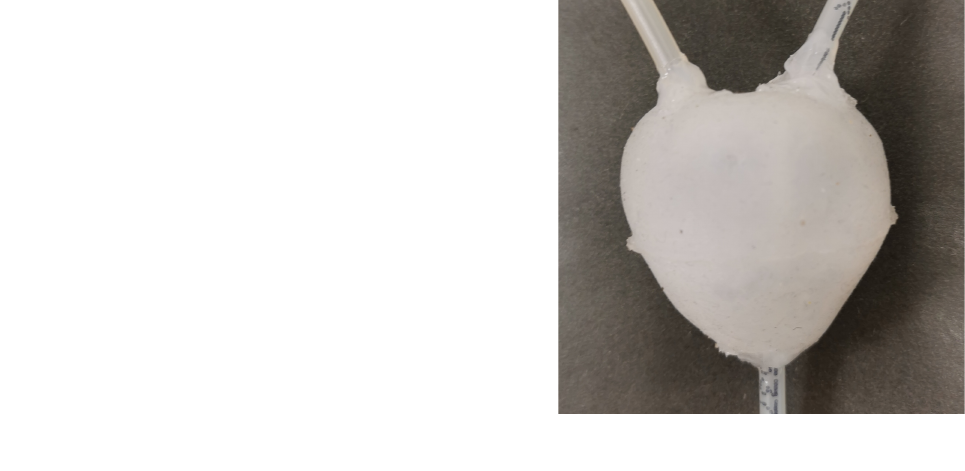
    \caption{Sensorized phantom: silicone body and finite-element method (FEM) model with three internal pneumatic chambers used as intrinsic pressure sensors. The phantom is fabricated by a lost-wax molding process using Ecoflex~00-10; full fabrication details are given in our previous work~\cite{tian_active_2025}.}%
    \label{fig:phantom_model}
\end{figure}

\subsection*{Compliance Matrix}
We generate a dataset by applying external forces in the three Cartesian directions
at 1000 surface locations sampled by farthest-point-sampling,
with FEM simulation in the Simulation Open Framework Architecture (SOFA)~\cite{coevoet_software_2017}. No internal actuation is applied,
so the response captures passive compliance. Each surface node is treated as a candidate sensor location;
for each sample we record node displacements and chamber pressures.

From the dataset, we construct the compliance matrix $S$ mapping input force changes $\text{d}F$ to sensor output changes $\text{d}Y$:
\begin{equation*}
    \text{d}Y = S \, \text{d}F.
\end{equation*}
The rows of $S$ are the $3 + 3N$ measurement channels
(3 chamber pressures plus 3 displacement components per surface node, with $N = 4052$);
the columns are the responses to the sampled unit-force inputs.
We then scale each modality (pressure, Pa; displacement, mm) so that its maximum sensitivity equals $1$,
mitigating their magnitude differences.

\subsection*{QR-Based Selection}

Following the pivoted-QR sensor selection of Manohar \textit{et al.}~\cite{manohar_data-driven_2018},
we initialize the basis with the three chamber pressure sensors.
At each iteration, we select the surface node $s^*$ whose three compliance rows are most linearly independent of those already chosen,
subject to a minimum-distance constraint:
\begin{equation*}
    \begin{aligned}
        s^* &= \operatorname*{argmax}_{i} \| R^{(k)}_i \|_F^2 \\
            &\quad \text{s.t.} \quad \min_{j \in \mathcal{S}} \text{dist}(p_i, p_j) \ge d_{\min}
    \end{aligned}
\end{equation*}
where $R^{(k)}_i$ is the residual of node~$i$'s compliance rows after orthogonal projection
against the rows of $\mathcal{S}$ at iteration~$k$,
$\| \cdot \|_F$ is the Frobenius norm,
and $d_{\min}$ is the minimum-distance threshold.

However, global QR treats all surface regions equally;
the resulting coverage is lowest on the
posterior DRE region (Figures~\ref{fig:reconstructability_qr_1} and~\ref{fig:reconstructability_qr_2}),
motivating a weighted variant that prioritizes a Region of Interest (ROI).

\subsection*{Weighted Greedy Search}
For a candidate sensor set $\mathcal{S}$, let $S^{(j)}_{\mathcal{S}} \in \mathbb{R}^{|\mathcal{S}| \times 3}$
denote the local sub-compliance matrix at surface node $j$:
its rows are the measurement channels of $\mathcal{S}$, and its three columns are the
unit-force responses at node $j$ in the $X$, $Y$, $Z$ directions.
We define the per-node Reconstructability Score
$\hat{s}_j(\mathcal{S}) = \sigma_{\min}\!\left(S^{(j)}_{\mathcal{S}}\right)$,
the minimum singular value of $S^{(j)}_{\mathcal{S}}$
(the E-optimality criterion~\cite{joshi_sensor_2009} applied per node);
a higher score means forces applied at node $j$ are more observable from $\mathcal{S}$.
We define the ROI as the posterior surface region within $15$\,mm of the expected DRE contact center.
Nodes inside the ROI receive weight $w_j = w_{\text{ROI}} = 5$; all others receive $w_j = 1$.
To keep the direct contact interface unobstructed, we forbid sensor placement inside the ROI.
Initialized with the three internal pressure sensors,
the algorithm iteratively appends the candidate that maximizes the weighted-mean reconstructability over the mesh,
subject to a minimum-distance constraint:
\begin{equation*}
    \begin{aligned}
        c^* &= \operatorname*{argmax}_{c \in \text{Cand}}
            \frac{\sum_{j} w_j \, \hat{s}_j(\mathcal{S} \cup \{c\})}{\sum_{j} w_j} \\
            &\quad \text{s.t.} \quad \min_{s \in \mathcal{S}} \text{dist}(c, s) \ge d_{\min}.
    \end{aligned}
\end{equation*}
where $\text{Cand}$ is the set of unselected surface nodes.
The algorithm selects $K = 10$ surface markers which, with the 3 internal pressure sensors,
form the 13-channel sensor set used throughout the remaining pipeline.

\section*{CONCLUSIONS AND DISCUSSION}

\begin{figure}[!t]
    \centering
    \subfloat[QR, posterior]{
        \includegraphics[width=0.2\textwidth]{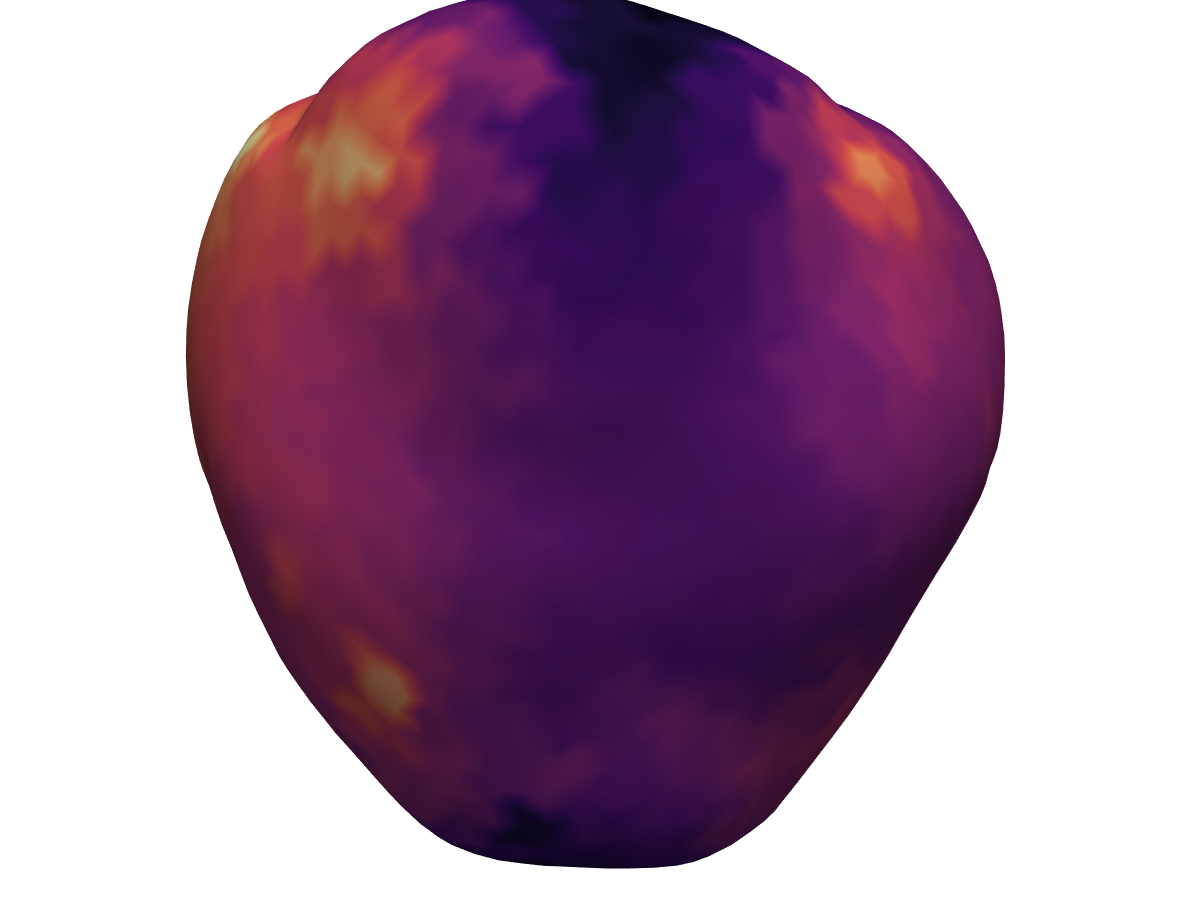}%
        \label{fig:reconstructability_qr_1}
    }
    \subfloat[QR, anterior]{
        \includegraphics[width=0.2\textwidth]{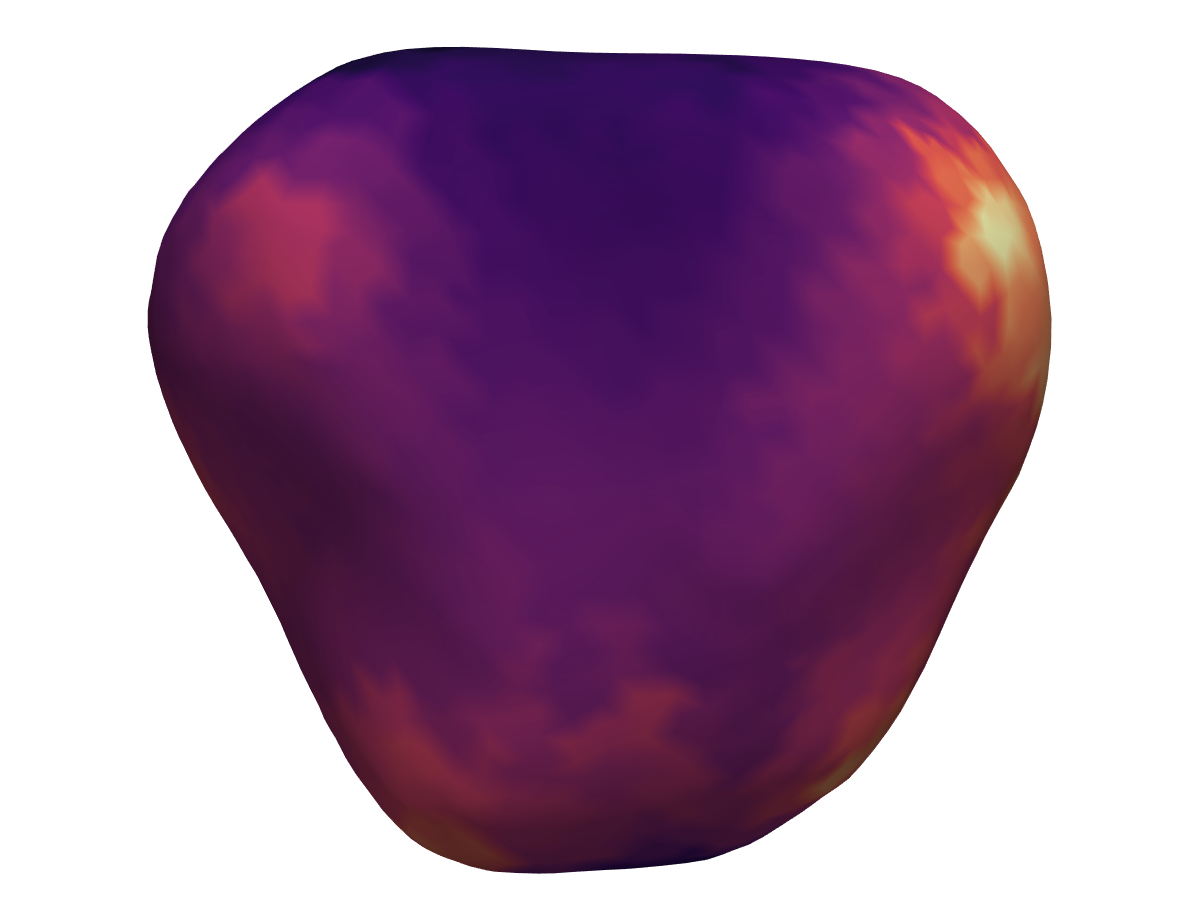}%
        \label{fig:reconstructability_qr_2}
    }
    \\
    \subfloat[Weighted greedy, posterior]{
        \includegraphics[width=0.2\textwidth]{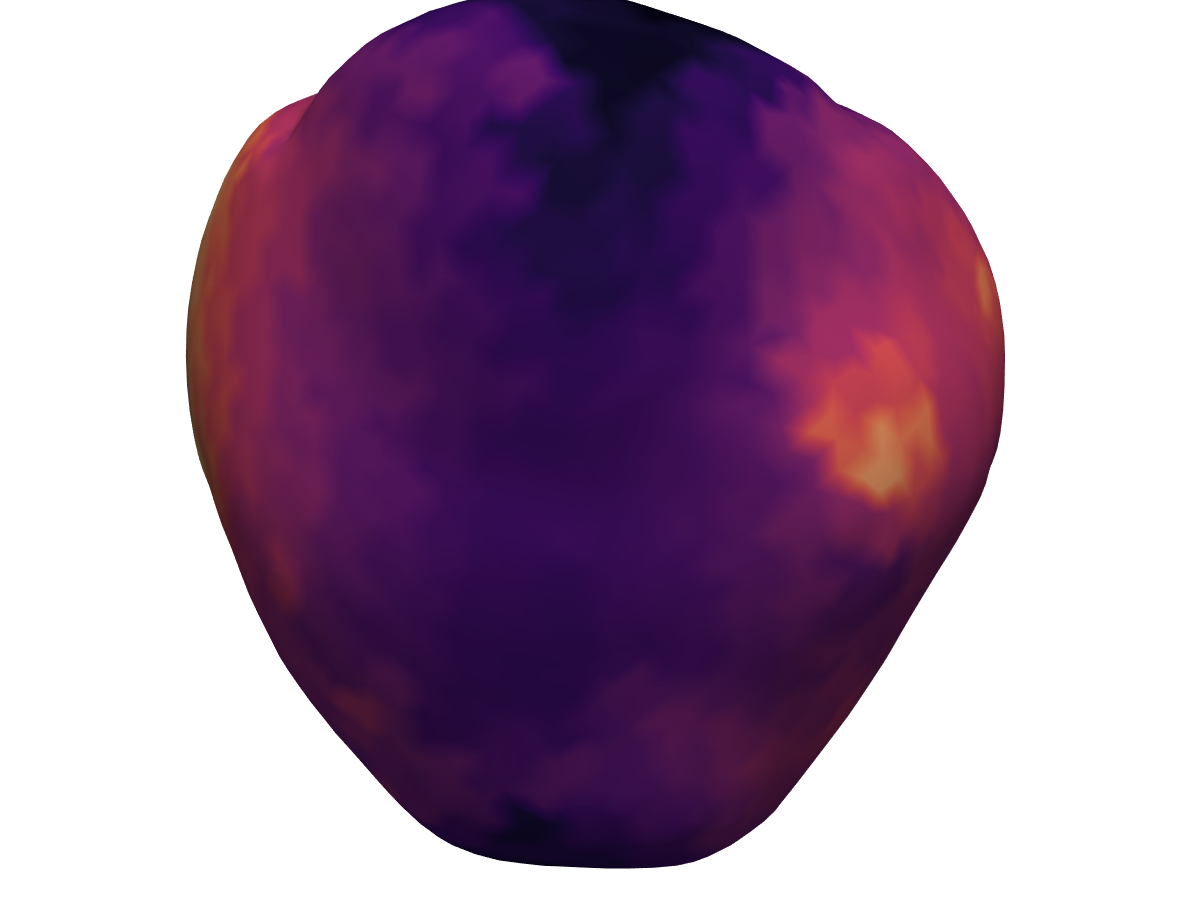}%
        \label{fig:reconstructability_weighted_greedy_1}
    }
    \subfloat[Weighted greedy, anterior]{
        \includegraphics[width=0.2\textwidth]{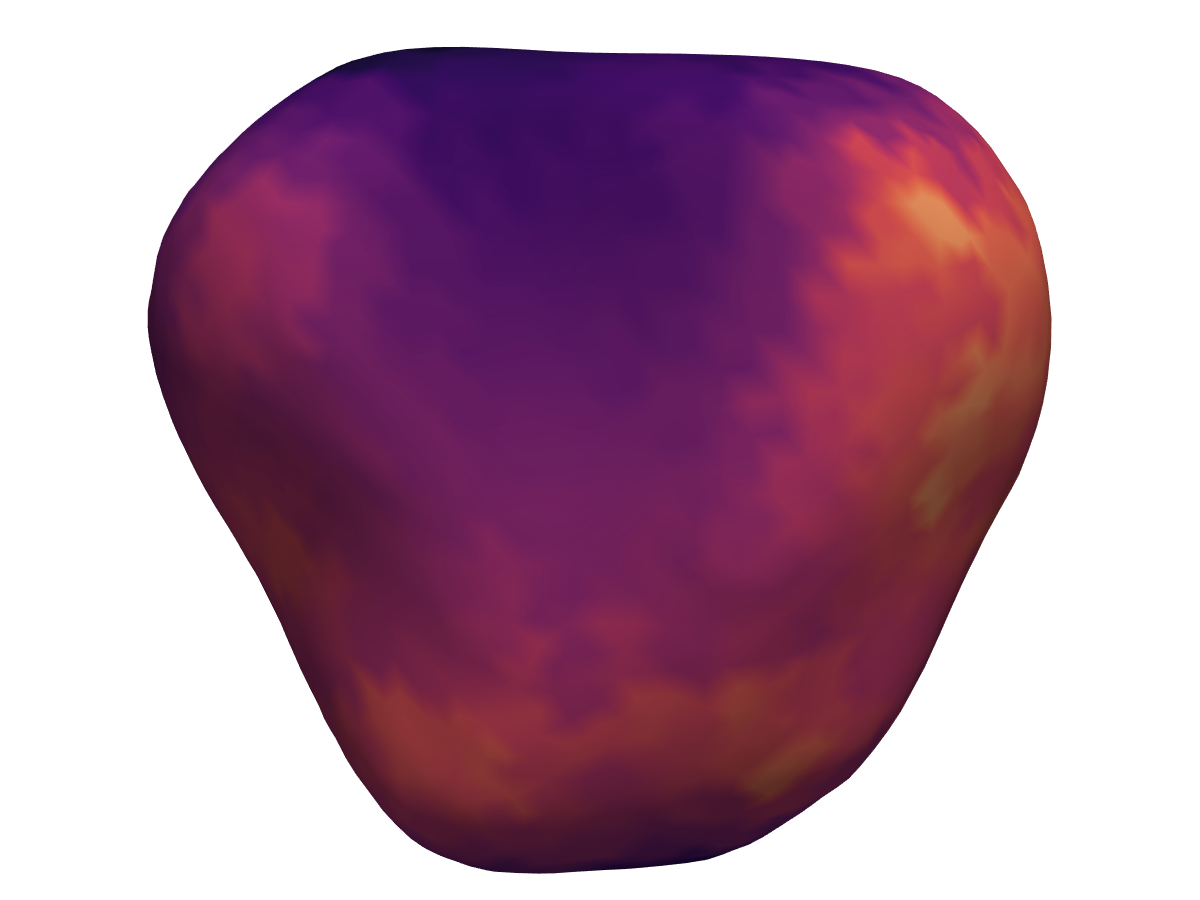}%
        \label{fig:reconstructability_weighted_greedy_2}
    }
    \\
    \includegraphics[width=0.42\textwidth]{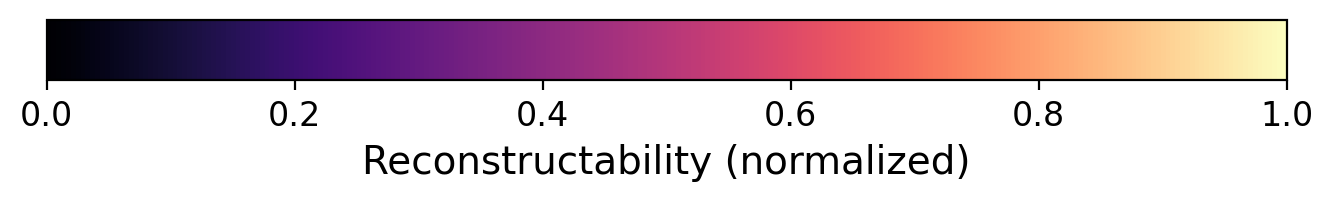}
    \caption{Reconstructability field with 10 markers selected by (a, b) QR and (c, d) weighted-greedy strategies, shown from posterior and anterior sides. Both fields are normalized by the maximum $\sigma_{\min}$ over surface nodes across both placements.}%
\end{figure}

We use 10 surface markers and 3 internal pressure sensors.
Figures~\ref{fig:reconstructability_qr_1} and~\ref{fig:reconstructability_qr_2} 
show the reconstructability field under QR-based placement: 
scores are high on the lateral surfaces but low on the posterior DRE region.
Under the proposed weighted-greedy placement,
which prioritizes the ROI in the selection score (Figures~\ref{fig:reconstructability_weighted_greedy_1} and~\ref{fig:reconstructability_weighted_greedy_2}),
observability concentrates in the ROI:
the mean of $\sigma_{\min}$ over ROI nodes
is 22.5\% higher under weighted-greedy than under QR.

\begin{figure}[!htbp]
    \centering
    \subfloat[Posterior view]{
        \includegraphics[width=0.3\columnwidth]{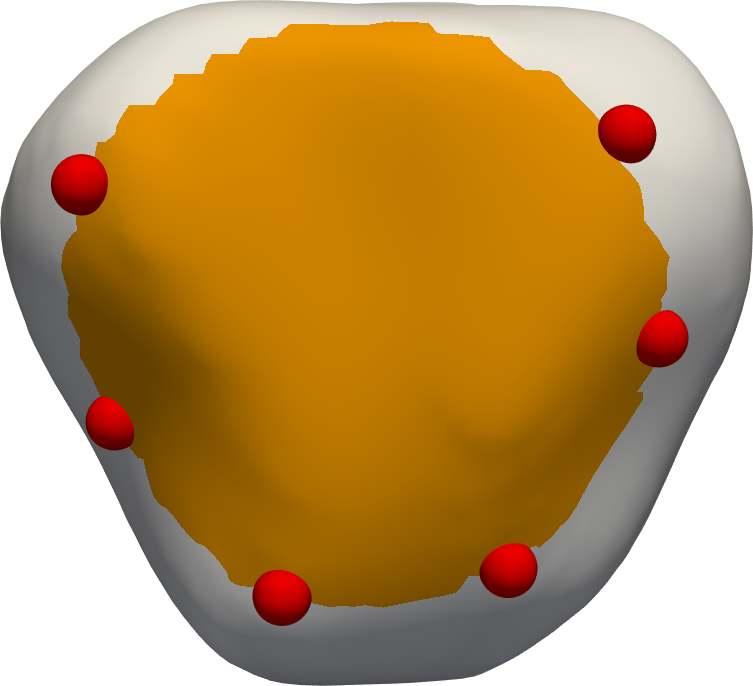}%
        \label{fig:sensor_config_post}
    }
    \hspace{0.05\columnwidth}
    \subfloat[Anterior view]{
        \includegraphics[width=0.3\columnwidth]{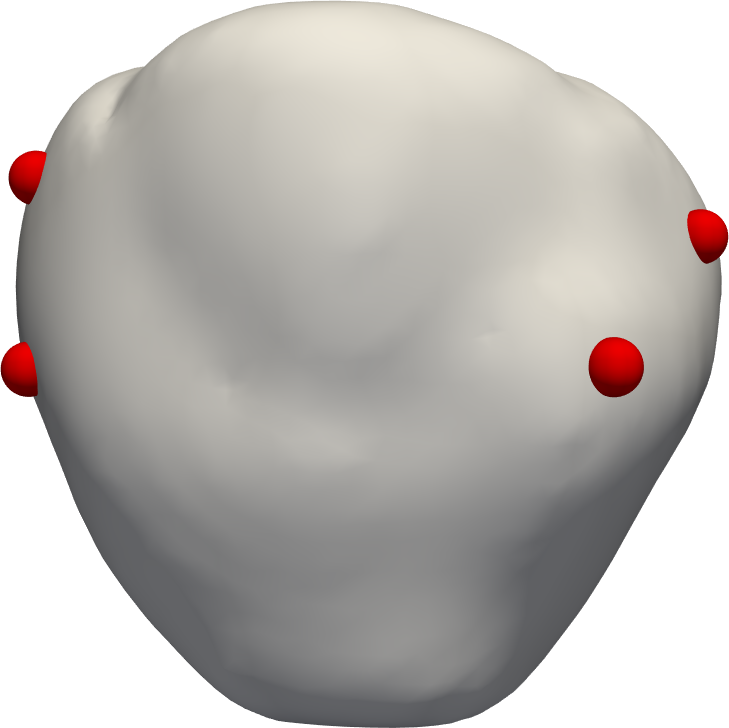}%
        \label{fig:sensor_config_ant}
    }
    \caption{Sensor configuration on the phantom. The ROI (orange patch, 15\,mm radius) covers the expected DRE contact zone; the 10 selected surface markers are shown in red.}%
    \label{fig:sensor_config}
\end{figure}

The selected configuration (Fig.~\ref{fig:sensor_config}) reflects the two mechanisms at work: 
the weighted score pulls markers toward the ROI where reconstructability matters most, 
while the exclusion zone forces them to ring the ROI boundary rather than lie on the contact interface itself. 
The remaining markers on the anterior side provide coverage for forces applied away from the DRE zone.

The current evaluation is in simulation.
Future work will validate the placement on hardware and on downstream tasks such as force localization,
extend the formulation to multi-ROI and moving-ROI scenarios,
and evaluate it on a broader set of soft-body geometries.

\bibliographystyle{IEEEtran}
\bibliography{ref}

\end{document}

%% file: prostate_model.pdf_tex
\begingroup%
  \makeatletter%
  \providecommand\color[2][]{%
    \errmessage{(Inkscape) Color is used for the text in Inkscape, but the package 'color.sty' is not loaded}%
    \renewcommand\color[2][]{}%
  }%
  \providecommand\transparent[1]{%
    \errmessage{(Inkscape) Transparency is used (non-zero) for the text in Inkscape, but the package 'transparent.sty' is not loaded}%
    \renewcommand\transparent[1]{}%
  }%
  \providecommand\rotatebox[2]{#2}%
  \newcommand*\fsize{\dimexpr\f@size pt\relax}%
  \newcommand*\lineheight[1]{\fontsize{\fsize}{#1\fsize}\selectfont}%
  \ifx\svgwidth\undefined%
    \setlength{\unitlength}{462.00299601bp}%
    \ifx\svgscale\undefined%
      \relax%
    \else%
      \setlength{\unitlength}{\unitlength * \real{\svgscale}}%
    \fi%
  \else%
    \setlength{\unitlength}{\svgwidth}%
  \fi%
  \global\let\svgwidth\undefined%
  \global\let\svgscale\undefined%
  \makeatother%
  \begin{picture}(1,0.48901622)%
    \lineheight{1}%
    \setlength\tabcolsep{0pt}%
    \put(0,0){\includegraphics[width=\unitlength,page=1]{prostate_model.pdf}}%
    \put(0.18250827,0.00925932){\color[rgb]{0,0,0}\makebox(0,0)[lt]{\smash{\begin{tabular}[t]{l}(a)\end{tabular}}}}%
    \put(0.77433283,0.00921597){\color[rgb]{0,0,0}\makebox(0,0)[lt]{\smash{\begin{tabular}[t]{l}(b)\end{tabular}}}}%
    \put(0,0){\includegraphics[width=\unitlength,page=2]{prostate_model.pdf}}%
    \put(0.425,0.41280431){\color[rgb]{0,0,0}\makebox(0,0)[lt]{\lineheight{1.60000002}\smash{\begin{tabular}[t]{l}Chamber 2\end{tabular}}}}%
    \put(0.425,0.28655586){\color[rgb]{0,0,0}\makebox(0,0)[lt]{\lineheight{1.60000002}\smash{\begin{tabular}[t]{l}Chamber 1\end{tabular}}}}%
    \put(0.425,0.15695111){\color[rgb]{0,0,0}\makebox(0,0)[lt]{\lineheight{1.60000002}\smash{\begin{tabular}[t]{l}Chamber 0\end{tabular}}}}%
    \put(0,0){\includegraphics[width=\unitlength,page=3]{prostate_model.pdf}}%
  \end{picture}%
\endgroup%